\pdfoutput=1

\documentclass[11pt]{article}

\usepackage[final]{acl}

\usepackage{times}
\usepackage{latexsym}
\usepackage{amsmath}
\usepackage{adjustbox}
\usepackage[T1]{fontenc}

\usepackage[utf8]{inputenc}

\usepackage{microtype}

\usepackage{inconsolata}

\usepackage{graphicx}
\usepackage{tabularx}
\usepackage{array}
\usepackage{booktabs}

\title{ESC-Judge: A Framework for Comparing Emotional Support Conversational Agents}

\author{Navid Madani \and Rohini Srihari \\
        Computer Science and Engineering - University at Buffalo \\
        Buffalo, NY, 14260 \\
        \texttt{\{smadani, rohini\}@buffalo.edu} }

\begin{document}
\maketitle
\begin{abstract}
Large Language Models (LLMs) increasingly power mental-health chatbots, yet the field still lacks a \emph{scalable}, \emph{theory-grounded} way to decide \emph{which} model is more effective to deploy.  
We present \textbf{ESC-Judge}, the first end-to-end evaluation framework that (i) \emph{grounds} head-to-head comparison of Emotional-Support LLMs (ES-LLMs) in an established psychological theory—Clara Hill’s \emph{Exploration–Insight–Action} (E-I-A) counselling model—thereby delivering a structured, interpretable lens on performance, and (ii) fully \emph{automates} the pipeline at scale.  
ESC-Judge proceeds in three stages:  
(1) it synthesizes realistic help-seeker roles by sampling empirically salient attributes (stressors, personality, life history);  
(2) it has two candidate ES-Agents conduct separate sessions with the \emph{same} role, isolating model-specific strategies; and  
(3) it asks a specialised judge LLM to issue pairwise preferences across rubric-anchored skills that exhaustively cover the E-I-A spectrum.  In our empirical study, ESC-Judge matches PhD-level annotators in \textbf{85\%} of Exploration, \textbf{83\%} of Insight, and \textbf{86\%} of Action decisions, demonstrating human-level reliability at a fraction of the cost. We release all code, prompts, synthetic roles, transcripts, and judgment scripts to catalyze transparent progress in emotionally supportive AI \footnote{\href{https://anonymous.4open.science/r/ESC-Judge-A508}{https://anonymous.4open.science/r/ESC-Judge-A508}}.

\end{abstract}

\begin{figure*}
    \centering
    \includegraphics[width=1\linewidth]{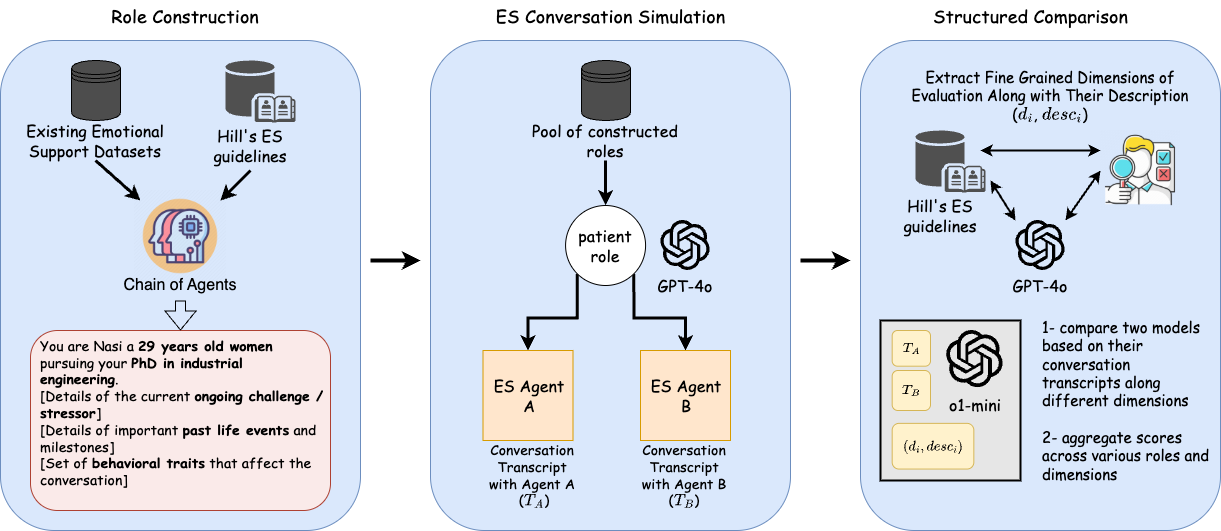}
    \caption{Overall pipeline of our proposed ES-Judge framework. Stage 1: constructs a diverse set of roles with various life backgrounds, demographics and behavioral attributes. Stage 2: conditioning on a fixed help seeker role, we have two emotional support (ES) models under test to participate in an emotional support conversation and we store the conversation transcripts. Stage 3: given carefully curated evaluation dimensions based on Hill's framework, we compare the capabilities of the two models under test on performing \emph{Exploration, Insight and Action.}}
    \label{fig:full-pip}
\end{figure*}

\section{Introduction}
\label{sec:introduction}

Large Language Models (LLMs) have begun powering mental-health chatbots and peer-support apps \citep{Stade2024LargeLM}.  
Because these agents interact with vulnerable users in high-stakes settings, the community urgently requires \emph{rigorous, theory-grounded evaluation} to decide which models are safe and effective to deploy. Most work still probes emotional-support quality with (i) \textbf{reference-based metrics} that score responses against a single gold transcript using lexical or semantic similarity measures such as BLEU, ROUGE or BERTScore and (ii) human annotation \citep{Zhao2023TransESCSE, Liu2021TowardsES, Zheng2023BuildingES}.  
Reference metrics demand large, professionally annotated corpora—expensive to create and culturally narrow—and implicitly assume one ``correct’’ reply, ignoring the multiplicity of valid counselling strategies.  
Similarity metrics reward paraphrase overlap while overlooking relational depth, empathic timing, and process adherence.  Finally, today’s best leaderboards still lean on \textbf{live human raters}; their judgements are slow, costly, subjective and often lack expert counselling knowledge, resulting in low inter-rater agreement and poor reproducibility.  

Clara Hill’s \emph{Exploration–Insight–Action (E-I-A)} framework offers an empirically validated lens on what \emph{ought} to happen in supportive dialogues \citep{Hill2014HelpingSF}. Yet, existing benchmarks neither operationalize this theory nor test models across the diverse personalities that modulate real conversations. Moreover, their reliance on continuous expert annotation prevents scaling beyond a few hundred pairs.  
A truly useful benchmark must therefore (i) encode counselling theory once, (ii) generalize to \emph{many} help-seeker personas, and (iii) scale to \emph{many} comparisons by being automated and not needing human intervention. This can enable scalable and self-supervised optimization of such agents in the future.

We introduce \textbf{ESC-Judge}, a three-stage, fully LLM-driven framework that addresses these gaps:

\begin{enumerate}
    \item \textbf{Help-seeker role construction}: We sample empirically influential traits (Big Five personality, coping style, trust level, social support, triggers)—all drawn from Hill’s text—to generate a spectrum of realistic help-seeker roles.
    \item \textbf{Emotional support conversation simulation}: Two candidate ES models converse independently with the \emph{same} help seeker role under identical generation settings, isolating model-specific strategies.
    \item \textbf{LLM Pairwise Judge}: A specialist judge model, instructed with the \emph{(E-I-A)} rubric, issues \texttt{A\;vs.\;B\;vs.\;tie} preferences for each fine-grained dimension. Pairwise comparison is cognitively easier than absolute scoring, avoids ad-hoc calibration, and aligns with real-world deployment choices.
\end{enumerate}

Our contributions can be summerized as follows:

\begin{itemize}
    \item \textbf{Theory-aligned benchmark}: First end-to-end emotional support judge pipeline grounded explicitly in Hill’s E-I-A counselling framework.  
    \item \textbf{Trait-driven realism}: Introduce personality-sensitive simulation that stress-tests ES agents across diverse user profiles.  
    \item \textbf{Scalable, expert-encoded judging}: Pairwise LLM judge achieves human-level reliability with match rate of \textbf{0.86, 0.85 and 0.83 on three categories of Exploration, Insight and Action}) while eliminating ongoing expert annotation costs.  
    \item \textbf{Open resources}: We release code, prompts, simulated roles, transcripts, and judgment scripts to catalyze transparent progress in supportive AI.
\end{itemize}

\paragraph{Road-map.}
Section~\ref{sec:related-work} reviews prior evaluation efforts; Section~\ref{sec:framework-overview} details ESC-Judge; Section~4 reports experiments; and Section~5 discusses limitations and future work.

\section{Related Work}
\label{sec:related-work}

\subsection{Human Evaluation of Emotional-Support Dialogues}

Early studies on emotional-support conversation (ESC) agents relied primarily on \textbf{human judgments}.  
The \textsc{ESConv} corpus \citep{Liu2021TowardsES} introduced theory-informed annotations of support strategies and evaluated systems via similarity measures between model utterances and human gold responses. Follow-up work continued to enlist either lay annotators or counseling experts to score generated dialogues for empathy, helpfulness, and coherence \citep{Zhao2024ESCEvalEE,  Zheng2022AugESCDA}. Although human evaluation captures nuanced relational qualities, it is expensive, yields only moderate inter-rater agreement for subjective traits, and scales poorly to the rapid iteration cycles of modern LLMs.

\subsection{Automated and LLM-Based Evaluation Protocols}

Given the limitations of manual annotation, researchers have explored \textbf{automatic metrics}. Standard lexical-overlap scores (BLEU, ROUGE, BERTScore) are the most common theme \citep{Liu2021TowardsES, Zhao2023TransESCSE, Zheng2023BuildingES}. Domain-specific proxies such as \emph{strategy following accuracy}—predicting whether a model follows the chosen strategy correctly—offer an alternative yet important aspect which is studied in \citep{Madani2024SteeringCL}.  

Recent advances turn large language models into \emph{reference-free judges}. Generic dialogue benchmarks like \citep{Dubois2024LengthControlledAA} and \citep{Zheng2023JudgingLW} prompt GPT-4 to conduct pairwise response comparisons and report moderate–high agreement with human preferences.

For the emotional-support domain, \citep{Zhao2024ESCEvalEE} combines role-played distressed users with multi-criteria human annotation and additionally trains a ranking model (ESC-Rank) to approximate expert scores. Despite using simulated roles, the constructed roles lack nuances that affect ES conversation and the ESC-Rank model is only trained on five utterances which is significantly short for assessing the full life-cycle of an emotional support conversation.

\section{ESC-Judge}

\subsection{Framework Overview}
\label{sec:framework-overview}

\noindent
\textbf{ESC-Judge} unfolds in three sequential stages that mirror a real-world counseling encounter while enforcing strict experimental control over the patient role, the evaluation rubric and the counselor characteristics.

\textbf{Stage 1 – Patient Role Construction.}
We construct a synthetic \emph{help-seeker} role by sampling a bundle of empirically salient client traits—e.g., ongoing stressors, important life events, big five personality traits and etc.  The resulting role prompt is injected into an instruction message that also defines session goals, ensuring every candidate model faces an identical, richly specified user role.

\textbf{Stage 2 – Simulating Emotional Support Conversation}
Each target Emotional-Support Agent \(\mathcal{M}_A\) and \(\mathcal{M}_B\) engages the simulated patient in an independent dialogue session of dynamic length. This design yields two parallel transcripts whose differences stem solely from the support strategies of the competing models.

\textbf{Stage 3 – LLM Judge Assessment.}
A specialized judge LLM receives the paired transcripts \((T_A, T_B)\) along with an evaluation dimension and outputs a preference—\(T_A \succ T_B\), \(T_B \succ T_A\), or \textsc{Tie} when neither response is clearly superior—along with rubric-anchored rationales that score each conversation across 9 fine-grained dimensions that represent Hill’s macro-dimensions (\textit{Exploration}, \textit{Insight}, \textit{Action}). Together, these three tightly coupled stages deliver a reproducible, plug-and-play testbed for head-to-head comparison of ES agents while remaining faithful to Clara Hill’s theoretical framework. Figure \ref{fig:full-pip} demonstrates the full pipeline of our proposed framework.

\begin{figure}
    \centering
    \includegraphics[width=\linewidth]{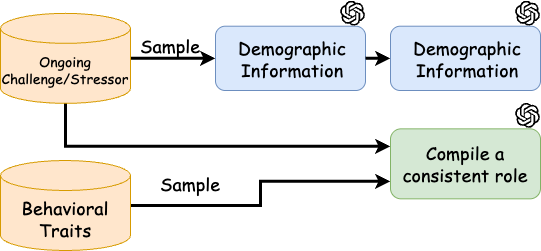}
    \caption{Role construction agents: Orange agents are random samplers based on pre-defined categories. Blue agents use generative prompts to explore the desired domain. The green agent only validates and compiles the final role without adding new information. Arrows represent the flow of data between agents.}
    \label{fig:role-construction}
\end{figure}

\subsection{Patient Role Construction}
\label{sec:patient-role}

While designing the role‐construction pipeline, we explicitly followed the factors that Hill identifies as most influential in an emotional-support encounter \cite{Hill2014HelpingSF}.  This stage implements a \emph{synthetic-role generation strategy} realised as a multi-step \textsc{Chain-of-Agents}: each agent, according to figure \ref{fig:role-construction} intracts with others to add a partial facet of information to enrich the proposed role.  The cascade halts with a consistency agent that reviews the full role to avoid inconsistent details. We explain each of these agents as follows.

\paragraph{Ongoing Challenge and Stressor.}
Because the presenting problem anchors the entire dialogue, we begin by sampling a salient life challenge from a curated pool, collated from existing emotional support and counseling datasets \cite{Liu2021TowardsES, Liu2023ChatCounselorAL}.6 Categories and 50 sub-categories are listed in Table~\ref{tab:stressors}.  Given a randomly chosen category \(c\), we uniformly sample a sub-category \(s\!\in\!c\) and send it to the next agent.

\paragraph{Demographic Information.}
This agent injects essential demographic descriptors to ground the role in a credible life context. We employ a generator prompt \cite{Chen2024GenQAGM} to yield a diverse yet consistent set of demographic attributes. Specifically, the agent adds gender, age, familial status, and occupation—all cross‑checked for coherence with the sampled stressor. For instance, when the stressor is \emph{Divorce or breakup}, familial status must reflect a dissolved partnership, whereas a \emph{retired veteran} profile is never paired with an eighteen‑year‑old. The result is a persona whose demographic identity harmonises with the ongoing challenge, promoting realistic downstream interactions. Details about the generator prompt and the configurations used, can be found in appendix \ref{sec:apx-role-demographic}.

\paragraph{Key Life Events.}
This agent imagines the help seeker’s personal history by generating a ranked list of $N$ salient life events spanning categories such as childhood trauma or positive experiences, family dynamics, romantic relationships, career milestones or failures, and loss or bereavement. Leveraging a nested generator prompt \citep{Chen2024GenQAGM}, it first explores a diverse set of candidate categories and then explores different scenarios within each category. It would then uniformly sample a subset to attach to the role. Prompts used for this section can be found in appendix \ref{sec:apx-role-keylife}.

\paragraph{Behavioral Traits}
According to Hill’s Helping Skills framework\,\cite{Hill2014HelpingSF}, a help-seeker’s behavioral profile can profoundly shape the course and effectiveness of an emotional-support dialogue.  Guided by the characteristics catalogued in the text, we organized the traits into 5 salient categories of 1) Big five personality traits 2) cognitive biases, thinking patterns and emotional baseline 3) response style towards therapist and trust in the process 4) social support network and coping mechanism and 5) triggers, sensitivities and self-soothing mechanisms each including some sub-categories. Afterwards, we sampled one representative variant from each to construct a concrete role for simulation.  Table~\ref{tab:trait-categories} lists the categories, dimensions, and exemplar variants used in our work. The selected variants along with a description are used to construct the role. More details can be found in appendix \ref{sec:apx-role-traits}.

Finally, a role construction agent takes the generated persona with demographics, key life events and sampled behavioral traits to construct a consistent full role. We utilize GPT-4o and langchain to construct the pipeline. Note that we have three types of agents in the pipeline as shown in Figure \ref{fig:role-construction}. Some agents only sample from a predefined data. Some are synthetic data generators and one is doing consistency check and re-writing. You can find details of each component, sample roles and prompts used for each agent in appendix \ref{sec:appendix-role-construction}.

\begin{table*}[t]
    \centering
    \footnotesize
    \begin{adjustbox}{width=0.95\linewidth}
        
    \begin{tabularx}{\textwidth}{@{}p{0.30\textwidth}X@{}}
        \toprule
        \textbf{Category} & \textbf{Sub-categories} \\
        \midrule
        Personal Loss \& Major Life Changes &
        Death of a loved one; Divorce or breakup; Family estrangement; Major illness or injury; Becoming a new parent; Caring for an aging family member; Pregnancy complications; Infertility or miscarriage; Social isolation; Immigration away from family \\\addlinespace
        Identity, Discrimination \& Social Challenges &
        Exploring LGBTQ+ identity; Lack of acceptance; Racial or gender discrimination; Workplace harassment; Identity crisis; Reputation damage \\\addlinespace
        Career \& Academic Pressures &
        Job loss; Toxic work environment; Career uncertainty; Burnout; Missed promotion; Academic failure; Completing a PhD; Job relocation; Fear of automation \\\addlinespace
        Financial \& Economic Stress &
        Significant debt; Inability to pay rent; Eviction; Medical bills; Loss of savings; Living paycheck-to-paycheck; Supporting dependents; Legal financial burdens; Bankruptcy \\\addlinespace
        Health \& Well-being &
        Chronic illness; Mental-health struggles; Sleep deprivation; Major surgery; Past trauma; Eating disorders; Addiction; Medication side-effects; Terminal illness \\\addlinespace
        Environmental \& Societal Stressors &
        Moving to a new country; Natural disasters; Political unrest or war; Victim of crime; Legal trouble; Forced lifestyle change (e.g., military service) \\
        \bottomrule
    \end{tabularx}
    \end{adjustbox}
    \caption{Stressors categories and sub-categories used during \emph{Ongoing Challenge and Stressor} sampling.}
    \label{tab:stressors}
\end{table*}

\subsection{Simulate Emotional Support Conversation}\label{subsec:simulation}

After constructing a diverse pool of patient roles (\ref{sec:patient-role}), we stage controlled dialogues to evaluate each emotional--support (ES) model under identical conditions. For every role~$r$ we create two conversations---$(r, ES_A)$ and $(r,ES_B)$---so that subsequent judgments compare model behaviour given the \emph{same} patient context. Dialogues are later scored against Hill's exploration--insight--action guidelines.

\paragraph{Dialogue engine}  The patient is realised as an autoregressive \emph{help seeker agent}: an LLM prompted with the role card plus the running history. The support agent is the ES model under test. Agents alternate turns with fixed generation settings (temperature 0.7, top--p 0.9, max 512 tokens to generate at each utterance) to isolate model--level differences.

\paragraph{Turn budget and early stopping}  LLM pairs often spiral into repetitious closing formalities (\emph{e.g.}, reciprocal thanks and farewells). To retain only the informative portion of the conversation, we (i) cap sessions at $T_{\max}=20$ turns and (ii) include a lightweight logistic regression end--of--conversation detector trained on 1K dialogues based on lexical 2-gram and 3-gram utterance features. When the model classifies the utterance as end--of--conversation, we stop the conversation at that point. More details about the training and evaluation of the end--of--conversation detector can be found in appendix \ref{sec:appendix-EOC}.

\subsection{LLM Judge Assessment}
\label{sec:judge-assessment}

\noindent
\textbf{Interactive rubric construction.}  
To transform Clara Hill’s three macro-chapters—\emph{Exploration}, \emph{Insight}, and \emph{Action}—into an operational scoring guide, we adopt a mixed LLM–human loop:

\begin{enumerate}
    \item \textbf{Chapter parsing.} We transform the book into markdown format and clean-up the resulting text. We use (\textsc{gpt\_4o}) and ingest each chapter and ask it to propose a candidate rubric: \emph{a proposed dimension, its definition and behavioural anchors}.
    \item \textbf{Author vetting.} Two authors independently screen the draft for faithfulness and specificity, merging identical dimensions and flagging vague ones (e.g., the initial ``\emph{exploration of feelings and thoughts}’’ was judged too broad. We split it into \emph{Encouragement of Emotional Expression} and \emph{Exploration of Thoughts and Narratives}).
    \item \textbf{LLM clarification rounds.} For every flagged item we prompt the model with the objection and request a sharper rewrite or removal. The loop typically converges in $\leq3$ rounds per chapter.
    \item \textbf{Pilot rating.} Annotators rate 50 dialogue pairs across all proposed dimensions with the provisional rubric; any item with agreement \(\kappa<0.5\) is re-phrased or discarded.
\end{enumerate}

The final rubric contains \textbf{9 fine-grained dimensions}:  
\emph{Exploration}: Encouragement of Emotional Expression, Exploration of Thoughts and Narratives, Empathic Understanding  
\emph{Insight}: Establish a Trusting Foundation, Assess Readiness for Insight, Use Gentle Challenges and Interpretations.  
\emph{Action}: Clarify the Desired Change, Ensure Readiness and Collaboration, Brainstorm and Evaluate Options.

Table \ref{tab:rubric-dimensions} categorizes all of these dimensions along with the definition of them.

\begin{table*}[t]
\footnotesize
\centering
\begin{adjustbox}{width=0.9\linewidth}
\begin{tabularx}{\textwidth}{>{\raggedright\arraybackslash}p{0.14\textwidth}|%
                        >{\raggedright\arraybackslash}p{0.27\textwidth}|%
                        X}
\textbf{Category} & \textbf{Dimension} & \textbf{Definition}\\\hline
Exploration & Empathic Understanding & Evaluate how well the model conveys a deep understanding of the user’s inner emotional world, reflecting feelings and aligning with the client’s subjective experience.\\ \hline
 & Encouragement of Emotional Expression & Determine if the model invites, explores, and validates emotional experiences—particularly helping the user articulate and tolerate difficult feelings.\\ \hline
 & Exploration of Thoughts and Narratives & Judge how well the model facilitates discussion of the user’s thoughts, beliefs, and personal stories through open‐ended questions and thoughtful restatements.\\\hline
Insight & Establish a Trusting Foundation & Create rapport and safety through empathic listening before offering deeper insights or interpretations.\\ \hline
 & Assess Readiness for Insight & Notice cues (e.g., confusion, ambivalence) that signal whether to probe deeper; avoid pushing insight if the user seems unready.\\ \hline
 & Use Gentle Challenges and Interpretations & Offer new perspectives tentatively, encouraging exploration of contradictions or underlying motives rather than dictating answers.\\ \hline
Action & Clarify the Desired Change & Invite exploration of the exact behaviour, situation, or decision the user wants to address, ensuring a specific goal before action planning.\\ \hline
 & Ensure Readiness and Collaboration & Check motivation to change and co‐create action plans, respecting self‐determination and context.\\ \hline
 & Brainstorm and Evaluate Options & Help generate multiple ideas, weigh feasibility, benefits, and challenges, and align options with values and needs.\\
 
\end{tabularx}
\end{adjustbox}
\caption{ESC‐Judge rubric dimensions and definitions.}
\label{tab:rubric-dimensions}
\end{table*}

\textbf{Pairwise judgement protocol.}  
For each dimension \(d\) we feed an \emph{o1-mini} judge with:

\begin{itemize}
    \item The full transcripts \((T_A,\,T_B)\)
    \item The plain-language definition of \(d\)
    \item A system instruction to (i) reason before judge and (ii) output a verdict: \texttt{A|B|tie}
\end{itemize}

We sample the judge twice with fixed temperature of 1.0 alternating the position of $T_A$ and $T_B$ to avoid position bias as emphasized by prior work \citep{Zheng2023JudgingLW}. If the verdicts change in these two sets of conditions, we choose \texttt{ tie} as the final verdict. If the output format does not match the prompted template (either the template is violated or the verdict is not given) we skip that instance. An example judge response is shown in appendix \ref{app:judge-prompts}.

\textbf{Aggregation of Judgements.}  
Let $w_{d}^{(A\!>\!B)}\in\{1,0,\tfrac12\}$ denote the outcome of model~A versus B on dimension $d$
\[
w_{d}^{(A\!>\!B)}=
\begin{cases}
1 & \text{judge prefers A}\\
0 & \text{judge prefers B}\\
\tfrac12 & \text{tie}
\end{cases}.
\]

\paragraph{Category–level comparison.}
For each Hill macro–category $c\in\{\textsc{Expl},\textsc{Ins},\textsc{Act}\}$ with dimension set $D_c$,  
the category score of A against B on a single \emph{role} $r$ is

\[
S_c^{(A\!>\!B)}(r)=\frac{1}{|D_c|}\sum_{d\in D_c} w_{d}^{(A\!>\!B)}(r).
\]

\paragraph{Across roles.}
Given a pool of simulated roles $\mathcal{R}$, we average:

\[
\bar S_c^{(A\!>\!B)}=\frac{1}{|\mathcal{R}|}\sum_{r\in\mathcal{R}} S_c^{(A\!>\!B)}(r).
\]

\paragraph{Decision rule.}
Model~A is judged \emph{preferred} to B in category $c$ if
\[
\bar S_c^{(A\!>\!B)}>\tfrac12 \quad\Longrightarrow\quad A \succ_c B,
\]
otherwise B is preferred; $\bar S_c^{(A\!>\!B)}=\tfrac12$ yields a tie.
We report preferences for each category separately rather than collapsing them into a single scalar, emphasising which stage of Hill’s framework drives overall superiority.

\section{Evaluation}

\subsection{Experimental Setup}

To evaluate the effectiveness of our judge framework, we conduct the following empirical study.  
First, we construct \textbf{25} patient roles, as described in Section~\ref{sec:patient-role}.  
We then assess three emotional–support agents: one proprietary model (GPT-4o-mini) and two open-source models (Llama-3.2-3B-Instruct and Llama-3.1-8B-Instruct).  
Each agent is prompted either \emph{with} or \emph{without} the general Hill guidelines (see Appendix~\ref{app:guidelines}), yielding six distinct agent configurations.

For the simulated help-seeker we use GPT-4o, conditioned on each constructed role.  
The help-seeker converses with a pair of support agents, and we record every dialogue as a triple
\((T_A, T_B, R_i)\), where \(T_A\) and \(T_B\) are the transcripts from agents \(A\) and \(B\), and \(R_i\) is the underlying patient role. Overally, this results in 375 triples.

Finally, our judge LLM (o1-mini reasoning) independently scores each transcript pair along the evaluation dimensions defined in Section~\ref{sec:judge-assessment}.  
These scores are aggregated into the \textsc{ESC-Judge} preference metric, following the procedure detailed in Section~\ref{sec:judge-assessment}.

\subsection{Does ESC-Judge distinguish between agents that do or do not follow Hill’s guidelines?}

\begin{figure*}
    \centering
    \includegraphics[width=1\linewidth]{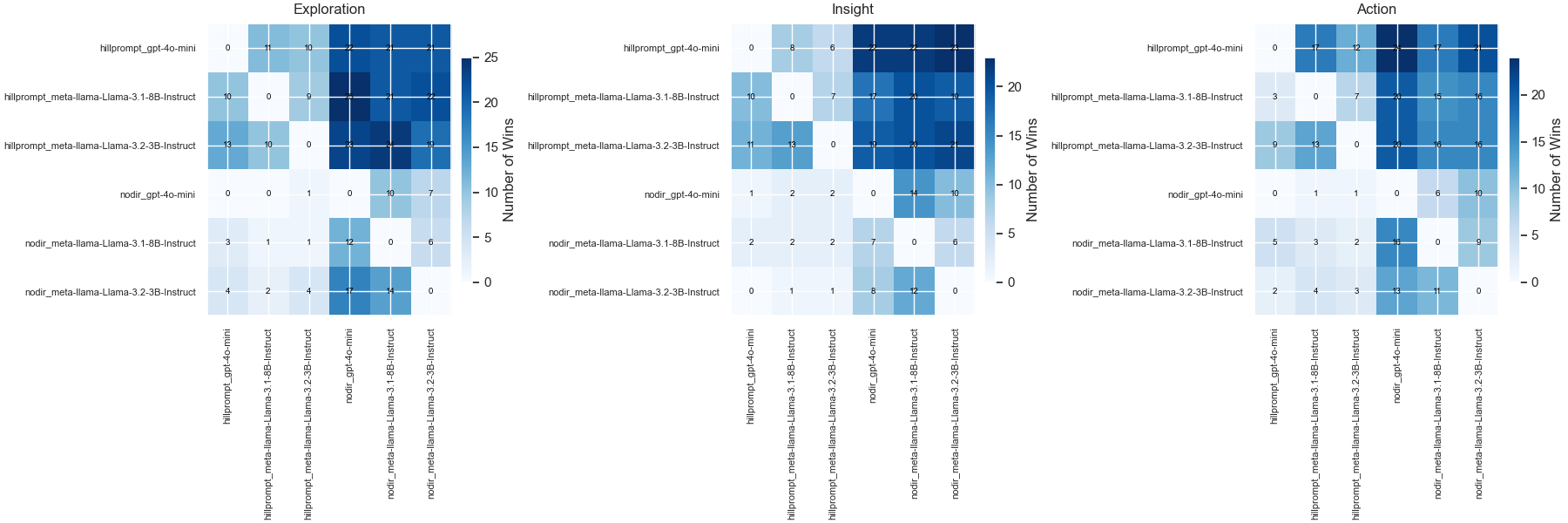}
    \caption{Comparison of the win-rate of different ES agents according to ESC-Judge framwork on three stages of exploration, insight and actoin.}
    \label{fig:eia-comp}
\end{figure*}

In our experimental setup, every emotional-support agent was prompted \emph{with} and \emph{without} Hill’s guidelines, producing two distinct agent groups—one explicitly aligned with the established directives. Figure~\ref{fig:eia-comp} presents pairwise comparisons of these agents across the three counselling stages: \emph{Exploration}, \emph{Insight}, and \emph{Action}. Winners are determined by the \emph{decision rule} described in Section~\ref{sec:judge-assessment}. 

\begin{figure*}[h]
    \centering
    \includegraphics[width=1\linewidth]{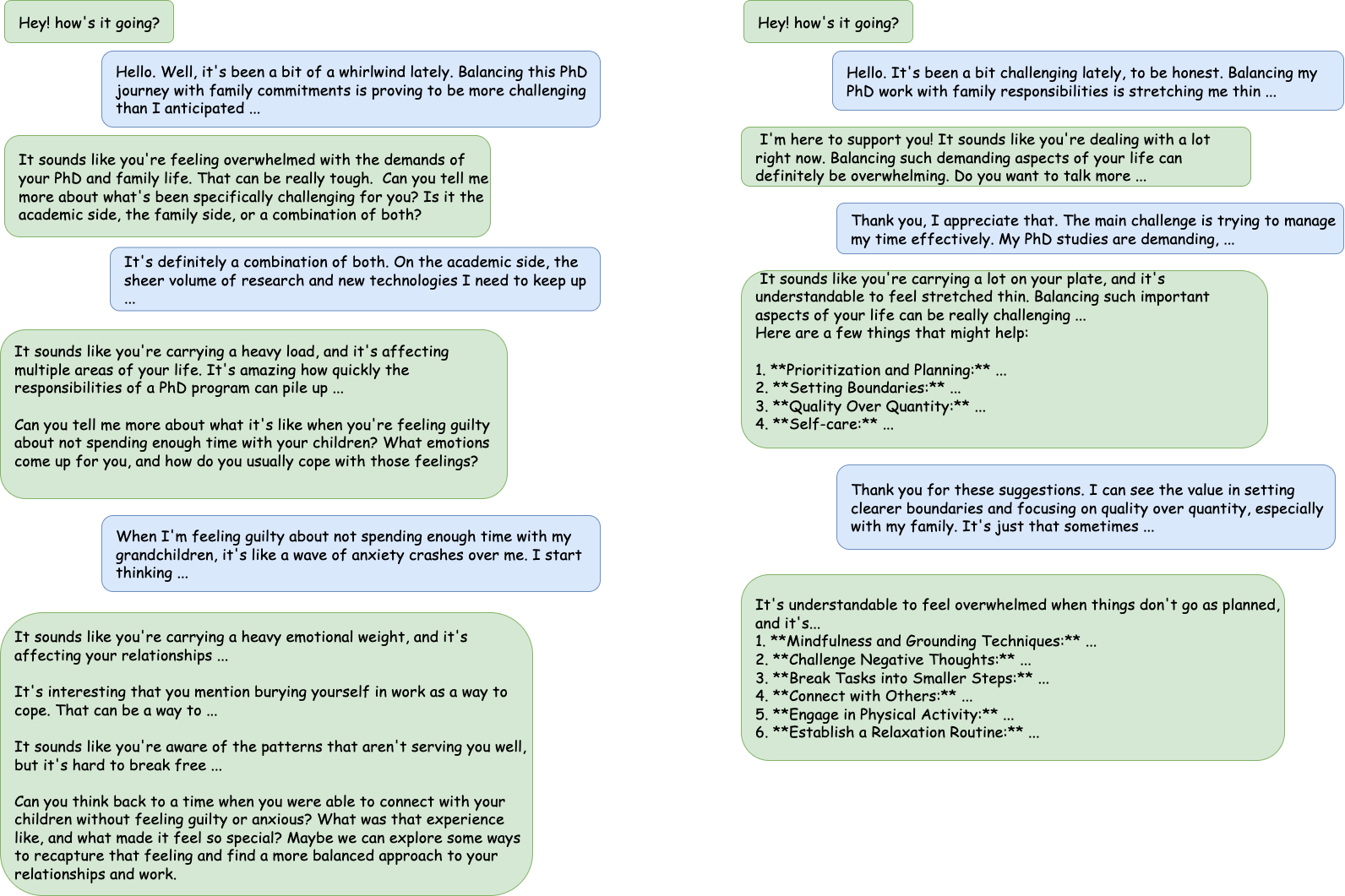}
    \caption{Left and right columns represent the first 7 turns of conversation between one help seeker role and two emotional support agents. One left ES agent is \emph{llama3.2-3b-instruct} with Hill's guideline prompt and on the right we have \emph{GPT-4o} without any guidelines as ES agent. \textbf{ESC-Judge} marks the left agent as the winner on \textbf{exploration} category.}
    \label{fig:exp-ex}
\end{figure*}

Across all three stages, agents instructed to follow Hill’s guidelines consistently outperform their uninstructed counterparts. The performance gap is smallest in the \emph{Action} stage, which aligns with the intuition that language models readily offer direct advice without structured guidance. Figure \ref{fig:exp-ex} shows an instance of comparison between two agents (only the first seven turns are shown) one with and the other without Hill's prompt. Our ESC-Judge marks the agent with Hill's guidelines as the winner on exploration category. It can be seen that this agent better demonstrates exploration stage and tries to elicit help seeker's emotions and thoughts instead of jumping to suggestions. 

\subsection{How well does ESC-Judge align with human annotators?}

To assess the reliability of \textsc{ESC-Judge}, we randomly sampled 100 conversation pairs and asked two PhD-level annotators to evaluate each pair across the same nine dimensions used by the judge. This produced \(100 \times 9 = 900\) human annotation instances. Following prior work, we consider only win–lose outcomes and discard ties when computing agreement.

Tables~\ref{tab:coarse-dim-match} and~\ref{tab:fine-dim-match} present the resulting match rates at both the coarse level (Exploration, Insight, Action) and the fine-grained dimension level. Counts differ between tables because some \textsc{ESC-Judge} outputs did not conform to the expected template and were removed during postprocessing. Appendix \ref{app:annotation}, demonstrates the annotation setup and the platform we used.

Aggregating \textsc{ESC-Judge} decisions as described in Section~\ref{sec:judge-assessment} yields a noticeably stronger correlation with human preferences. We apply the same aggregation procedure to the human annotations, retaining only win–lose cases, and then compute the match rate for each coarse category. As shown in Table~\ref{tab:aggregated-match}, \textsc{ESC-Judge} aligns with human judgments in \textbf{86\%, 83\% and 85\%} of cases for \emph{Exploration}, \emph{Insight}, and \emph{Action} respectively.

\begin{table}[t]
\centering
\small
\begin{tabular}{lcc}
\toprule
\textbf{Coarse Dimension} & \textbf{Match Rate} & \textbf{Count} \\
\midrule
Action        & 0.851852 & 27 \\
Exploration   & 0.857143 & 28 \\
Insight       & 0.827586 & 29 \\
\bottomrule
\end{tabular}
\caption{Aggregated match rates and counts for each coarse‐grained dimension.}
\label{tab:aggregated-match}
\end{table}

\begin{table}[t]
\centering
\small
\begin{adjustbox}{width=\columnwidth}
\begin{tabular}{lcc}
\toprule
\textbf{Fine-grained Dimension} & \textbf{Match Rate} & \textbf{Count} \\
\midrule
Assess Readiness for Insight                     & 0.577465 & 71 \\
Brainstorm and Evaluate Options                  & 0.717647 & 85 \\
Clarify the Desired Change                       & 0.753247 & 77 \\
Empathic Understanding                           & 0.911392 & 79 \\
Encouragement of Emotional Expression            & 0.861111 & 72 \\
Ensure Readiness and Collaboration               & 0.771084 & 83 \\
Establish a Trusting Foundation                  & 0.835616 & 73 \\
Exploration of Thoughts and Narratives           & 0.860759 & 79 \\
Use Gentle Challenges and Interpretations        & 0.761364 & 88 \\
\bottomrule
\end{tabular}
\end{adjustbox}
\caption{Match rates and counts for each fine-grained dimension.}
\label{tab:fine-dim-match}
\end{table}

\begin{table}[t]
\centering
\small
\begin{tabular}{lcc}
\toprule
\textbf{Coarse Dimension} & \textbf{Match Rate} & \textbf{Count} \\
\midrule
Action        & 0.739130 & 322 \\
Exploration   & 0.878261 & 230 \\
Insight       & 0.727273 & 242 \\
\bottomrule
\end{tabular}
\caption{Match rates and counts for each coarse‐grained dimension.}
\label{tab:coarse-dim-match}
\end{table}

\section*{Limitations}

While \textsc{ESC‐Judge} advances automated, theory-grounded comparison of emotional-support agents, several important limitations remain:

\paragraph{Personality and trait coverage.}
Our role–construction pipeline samples from a finite catalogue of stressors, demographic profiles, and behavioural traits drawn from Clara Hill’s framework and related datasets.
Although the resulting roles span many salient factors, they cannot exhaust the full spectrum of human personalities, cultural backgrounds, or situational nuances encountered in practice.
Deployments in new domains should therefore augment the role pool—or collect real user data—to ensure adequate representativeness. In addition this work is only considering a single establish theory while there are many other approaches and framework in emotional support that can be studied.

\paragraph{Need for expert dialogue review.}
The judge model evaluates transcripts \emph{post hoc}; it does not interactively probe follow-up questions or verify factual accuracy during the conversation.
Before clinical or large-scale deployment, candidate systems should be vetted through live sessions with trained mental-health professionals to catch subtleties—such as misinterpretation of client affect or inappropriate self-disclosure—that the automated rubric may overlook.

\paragraph{Safety and regulatory compliance.}
We assess counselling quality but do not perform a thorough safety audit.
Models may still produce harmful advice, hallucinate clinical facts, or violate jurisdiction-specific regulations (e.g., HIPAA, GDPR).
Comprehensive red-team testing, toxicity filtering, and legal review are essential prerequisites for any real-world rollout.

\paragraph{Language scope.}
All experiments are conducted in English with largely Western cultural assumptions embedded in both the role prompts and the Hill-based rubric.
Performance may degrade for other languages or cultural contexts where concepts of emotional expression and counselling norms differ.
Future work should translate and culturally adapt the rubric, then replicate our study in multilingual settings.

\paragraph{Evaluation scale and stability.}
Although pairwise judging reduces variance compared to absolute scoring, we rely on a single small reasoning model (\texttt{o1-mini}) and sample each comparison only twice.
Larger judges, more sampling, and cross-model ensembling could further stabilise decisions—especially on fine-grained dimensions where current human alignment still falls below perfect agreement.

Taken together, these limitations highlight that \textsc{ESC‐Judge} is best viewed as a \emph{research benchmark} rather than a deployment-ready certification tool; practitioners must combine it with extensive human expert testing, safety analysis, and cultural adaptation before trusting ES-LLMs in sensitive real-world scenarios.

\bibliography{acl_latex}

\appendix

\section{Role Construction}
\label{sec:appendix-role-construction}

\begin{figure*}
    \centering
    \includegraphics[width=1\linewidth]{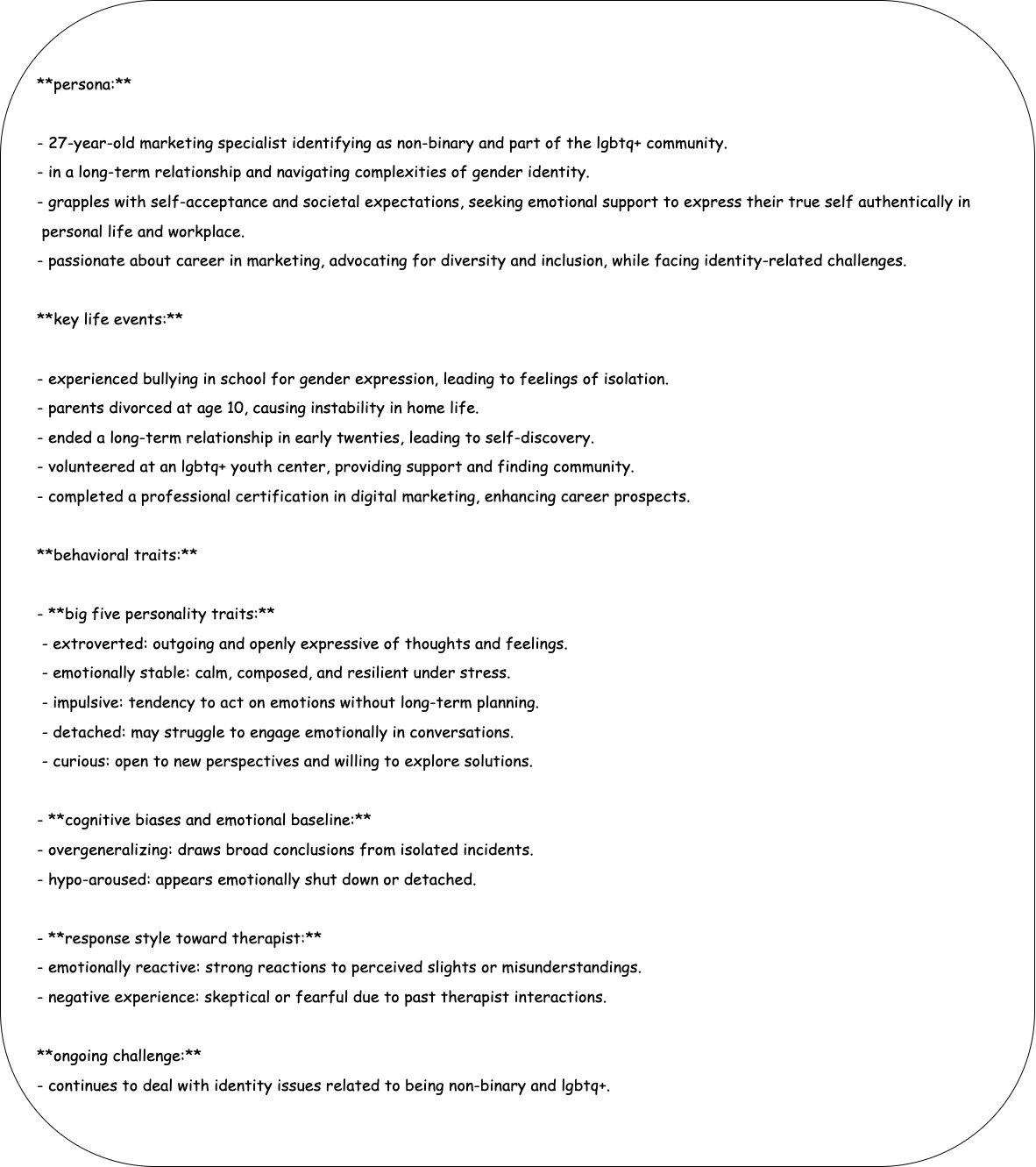}
    \caption{A full sample role from the role construction pipeline}
    \label{fig:sample-role}
\end{figure*}

Figure \ref{fig:sample-role} demonstrates an example finalized role, out of the \emph{role construction} agentic pipeline.

\subsection{Demographic Information}
\label{sec:apx-role-demographic}

\begin{figure*}
    \centering
    \includegraphics[width=1\linewidth]{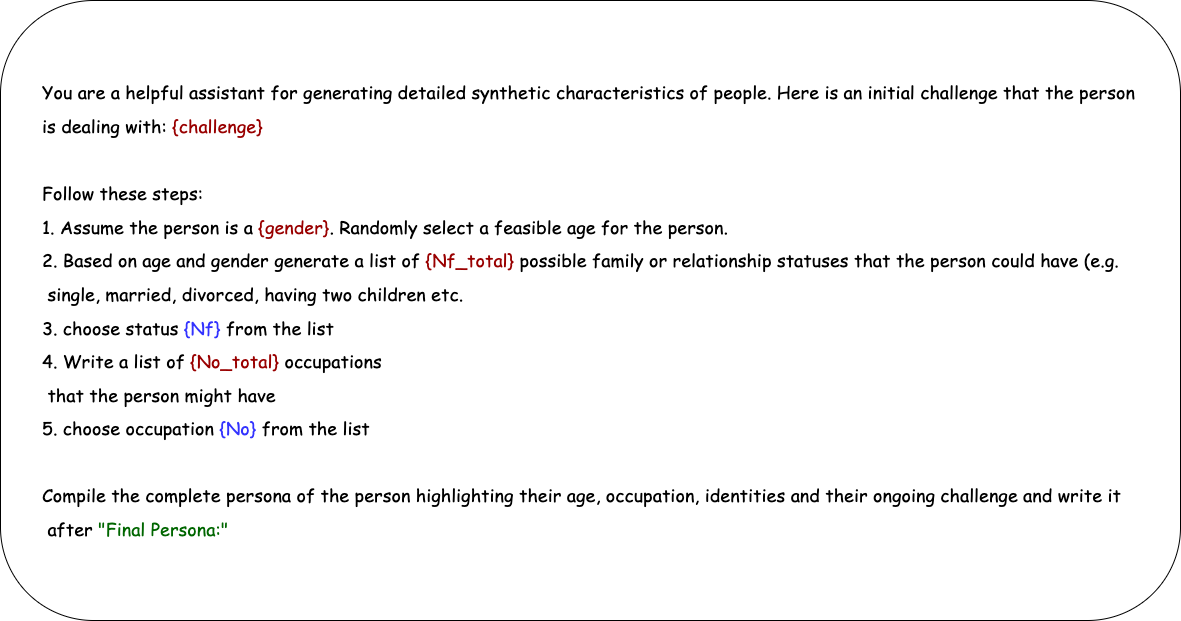}
    \caption{Generator prompt used for demographic information agent.}
    \label{fig:dem-prompt}
\end{figure*}

In this section, we use a generative prompt as explained in \citep{Chen2024GenQAGM}. Figure \ref{fig:dem-prompt} shows the prompt we used for this agent. For this prompt, we feed the information provided in curly brackets. \emph{challenge} is given from a previous agent, \emph{gender} is sampled from the set of \{man, woman\}, \emph{Nf\_total} is a configurable parameter that we set to 5, \emph{No\_total} is set to 10, \emph{Nf} and \emph{No} are randomly and uniformly sampled from 1 to 5 and 10 respectively. This way, the model explores a list of possible candidates and uniformly chooses one at each generation step.

\subsection{Key Life Events}
\label{sec:apx-role-keylife}

\begin{figure*}
    \centering
    \includegraphics[width=1\linewidth]{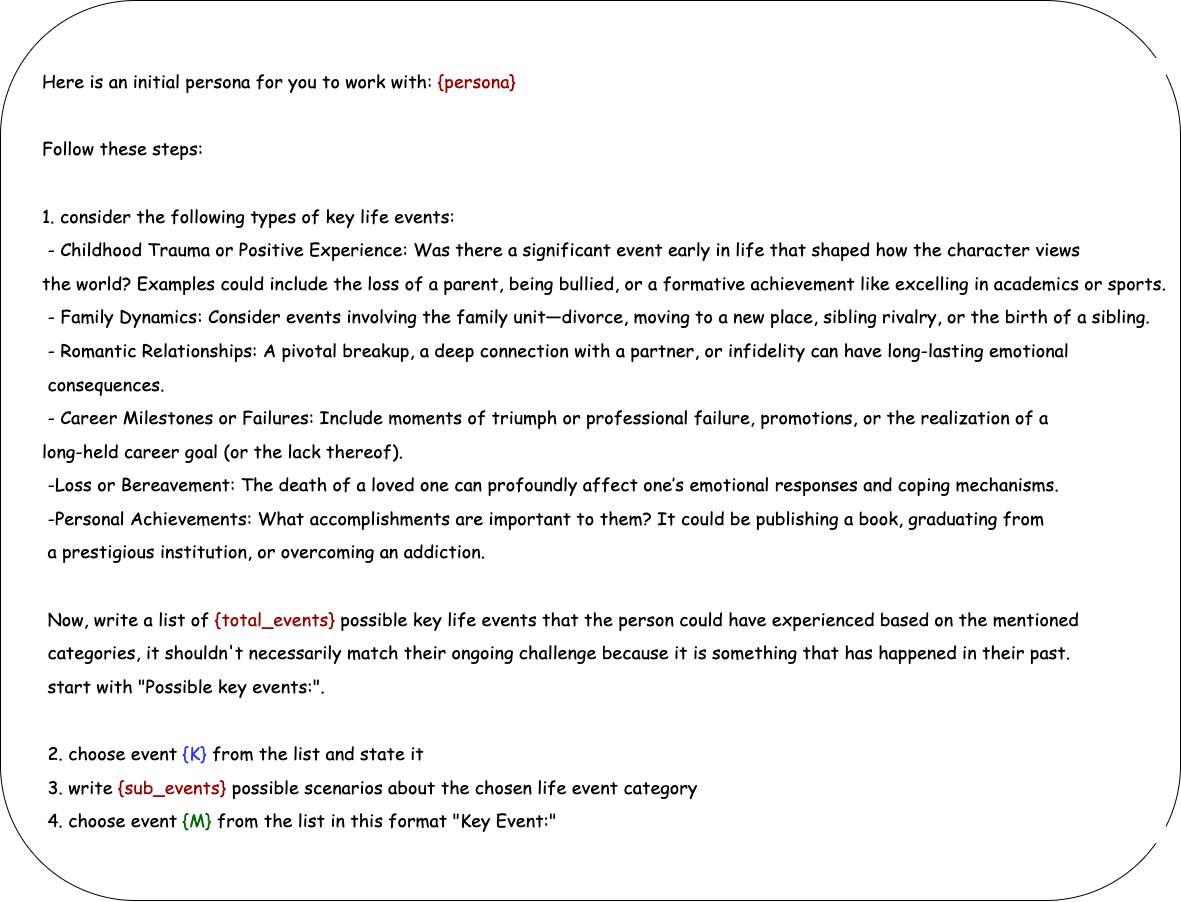}
    \caption{Nested generator prompt used for \emph{key life events} generator agent}
    \label{fig:keylife-prompt}
\end{figure*}

For building key life events, we use a nested generative prompt to better explore the domain of possible options. Figure \ref{fig:keylife-prompt} shows the prompt that we used for this agent. \emph{persona} is given from the previous agent (demographic information agent), then a list of examples is provided to the prompt to guide it to write $total\_events=20$ events. Then the agent chooses \emph{K}th element of the list randomly choosen from 1 to 20. Afterwards the agent is forced to write $sub\_events=25$ scenarios within that category of events and randomly choose \emph{M}th element. This way the agent explores a taxonomy of possible events and chooses one randomly. We repeat this process randomly between 1 and 4 times for each role, to generate between 1 and 4 scenarios for the key life events part.

\subsection{Behavioral Traits}
\label{sec:apx-role-traits}

We identify five overarching categories comprising thirteen sub-categories of help-seeker behavioral traits that, according to Hill’s textbook \citep{Hill2014HelpingSF}, meaningfully shape the course of an emotional-support conversation. The categories, their sub-categories, and the available variant options are summarized in Table~\ref{tab:trait-categories}. Brief descriptions of each variant are provided in Table~\ref{tab:trait-desc}. During role construction, the \emph{behavioral-traits} agent samples one variant from each category and forwards the selected variants—with their accompanying descriptions—to the next agent.

\begin{table*}[htbp]
\small
\centering
\begin{adjustbox}{width=0.95\linewidth}
    
\begin{tabularx}{\textwidth}{>{\raggedright\arraybackslash}p{0.30\textwidth}|%
                        >{\raggedright\arraybackslash}p{0.32\textwidth}|%
                        >{\raggedright\arraybackslash}p{0.32\textwidth}}
                        \hline
\textbf{Category} & \textbf{Sub-category} & \textbf{Trait Options}\\\hline
Big Five Personality Traits & Extraversion & Introverted\\
 &  & Extroverted\\ \hline
 & Neuroticism (Emotional Stability) & Emotionally Stable\\
 &  & Emotionally Reactive\\ \hline
 & Conscientiousness & Disciplined\\
 &  & Impulsive\\ \hline
 & Agreeableness & Empathetic\\
 &  & Detached\\  \hline
 & Openness to Experience & Curious\\
 &  & Traditional\\\hline
Cognitive Biases, Thinking Patterns, and Emotional Baseline & Cognitive Biases & Catastrophizing\\
 &  & Black and white thinking\\
 &  & Overgeneralizing\\
 &  & Emotional reasoning\\ \hline
 & Emotional Baseline & Hyper-aroused\\
 &  & Hypo-aroused\\
 &  & Emotionally volatile\\\hline
Response Style Toward the Therapist and Trust in the Process & Response Style & Easily reassured\\
 &  & Needs logical explanation\\
 &  & Resistant and defensive\\
 &  & Emotionally reactive\\ \hline
 & Trust in the Process & Positive experience\\
 &  & Negative experience\\
 &  & First-time experience\\\hline
Social Support Network and Coping Mechanisms & Social Support Network & Strong support\\
 &  & Weak or nonexistent support\\
 &  & Conflicted support\\ \hline
 & Coping Mechanisms & Adaptive coping\\
 &  & Maladaptive coping\\
 &  & Avoidant coping\\\hline
Triggers, Sensitivities, and Self-soothing Mechanisms & Triggers & Topic-specific triggers\\
 &  & Therapist-specific triggers\\
 &  & Environmental triggers\\ \hline
 & Self-soothing Mechanisms & Rationalization\\
 &  & Distraction\\
 &  & Suppression\\ \hline
\end{tabularx}
\end{adjustbox}
\caption{Hierarchy of simulated help seeker behavioral traits}
\label{tab:trait-categories}
\end{table*}

\begin{table*}[htbp]
\small
\centering
\begin{adjustbox}{width=0.95\linewidth}
\begin{tabularx}{\textwidth}{>{\raggedright\arraybackslash}p{0.33\textwidth}|%
                        >{\raggedright\arraybackslash}X}
\textbf{Variant} & \textbf{Description}\\\hline
Introverted & You are more reserved and may need more prompting to share thoughts and emotions.\\
Extroverted & You are outgoing and engages openly, easily expressing thoughts and feelings.\\
Emotionally Stable & You remain calm and composed, handling stress with resilience.\\
Emotionally Reactive & You experience heightened emotional responses, struggling with anxiety or mood swings.\\
Disciplined & You are goal-oriented, organized, and methodical in addressing their concerns.\\
Impulsive & You struggle with planning and may act on emotions without considering long-term consequences.\\
Empathetic & You are warm, trusting, and open to collaboration in the helping process.\\
Detached & You may be skeptical, resistant, or struggle to engage emotionally in conversations.\\
Curious & You are open to new perspectives, willing to explore different solutions and reflect on emotions.\\
Traditional & You prefer familiar approaches, may resist change, and values structured, predictable guidance.\\\hline
Catastrophizing & You expect the worst possible outcome in every situation.\\
Black-and-white thinking & You view situations as all good or all bad, with no middle ground.\\
Overgeneralizing & You make broad conclusions based on isolated incidents.\\
Emotional reasoning & You believe that their emotions reflect objective reality (e.g., feeling worthless means they are worthless).\\
Hyper-aroused & You are restless, easily triggered, and may have difficulty focusing due to heightened anxiety.\\
Hypo-aroused & You appear emotionally shut down or detached, showing little emotional engagement.\\
Emotionally volatile & You experience rapid emotional swings, moving between different emotional states quickly.\\\hline
Easily reassured & You calm down quickly with reassurance, validation, or soothing techniques.\\
Needs logical explanation & You respond best to structured, evidence-based interventions and logical reasoning.\\
Resistant and defensive & You are skeptical of the therapist, may challenge suggestions, and is resistant to intervention.\\
Emotionally reactive & You react strongly to perceived slights or misunderstandings, possibly becoming angry or withdrawn.\\
Positive experience & You trust the therapist and the process based on prior success.\\
Negative experience & You are skeptical or fearful of the process due to past negative interactions with therapists.\\
First-time experience & You are unfamiliar with therapy but open to exploring it, though they may be apprehensive.\\\hline
Strong support & You have a reliable network of family and friends for emotional support, which can help or hinder progress.\\
Weak or nonexistent support & You feel isolated and may rely heavily on the therapist for emotional regulation.\\
Conflicted support & You have strained relationships with key people in their life, potentially increasing stress.\\
Adaptive coping & You use healthy coping strategies like mindfulness, exercise, or seeking social support.\\
Maladaptive coping & You engage in destructive coping strategies such as substance abuse or aggression.\\
Avoidant coping & You avoid confronting painful issues by deflecting or minimizing the problem.\\\hline
Topic-specific triggers & Certain subjects, such as family or past trauma, provoke a strong emotional response from the client.\\
Therapist-specific triggers & The therapist’s tone, body language, or choice of words may unintentionally set off a negative reaction.\\
Environmental triggers & External factors such as background noise or discomfort in the setting may distract or distress the client.\\
Rationalization & You try to calm themselves by using logic to downplay emotional distress.\\
Distraction & You shift focus away from anxiety by talking about unrelated subjects or asking unrelated questions.\\
Suppression & You ignore or suppress emotions, which may lead to delayed or intensified emotional reactions later.\\
\end{tabularx}
\end{adjustbox}

\caption{Variant–description mapping for help seeker behavioral traits}
\label{tab:trait-desc}
\end{table*}

\section{Emotional Support Conversation Simulation}
\label{sec:appendix-simulation}

\subsection{End of Conversation Detection}
\label{sec:appendix-EOC}

\paragraph{End-of-conversation detector.}
We train an end-of-conversation (EoC) classifier via weak supervision. Starting with the complete set of simulated dialogues generated by all agents, we split the data 80 / 20 into train and test partitions. Each instance is formed from two consecutive utterances, which we vectorise with a TF–IDF model (\texttt{scikit-learn}) using uni-, bi-, and trigrams, removing English stop-words and discarding terms with a document frequency above 0.4. We manually label the test instances.

Weak labels are assigned to training examples as follows: an example is marked 1 (EoC) only if the dialogue has more than six turns \textbf{and} at least one farewell phrase from the list below appears; otherwise it is labeled 0.
\textit{“Take care, and talk soon”, “Good bye”, “I look forward to our next conversation”, “See you later”, “Take care”, “Bye for now”, “Catch you later”, “See you soon”, “Talk to you later”, “It was nice talking to you”, “See ya”, “Until next time”, “bye”, “see you”, “Good night”, “Farewell”, “Have a great day”, “Thanks, that’s all”, “That’s it, thanks”}.

We fit a logistic-regression classifier on this weakly labeled training set. On the held-out test split, the model achieves 0.91 accuracy and an F$_1$ score of 0.81. Importantly, recall for \emph{non-EoC} instances is 0.99, ensuring we terminate conversations only when highly confident. Recall for EoC instances is 0.70, so about 30 \% of true endings are missed—occasions in which agents may continue polite formalities until the turn budget is reached or a later detection fires.

\label{app:guidelines}
\section{Emotional Support Conversation LLM Prompts}

\begin{figure*}
    \centering
    \includegraphics[width=1\linewidth]{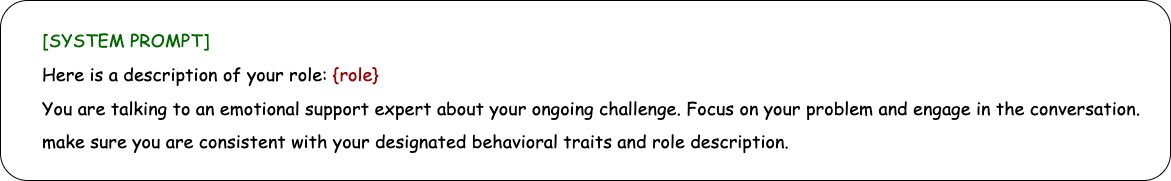}

    \caption{Prompt template used for the help seeker LLM}
        \label{fig:seeker-prompt}
\end{figure*}

\begin{figure*}
        \centering
        \includegraphics[width=1\linewidth]{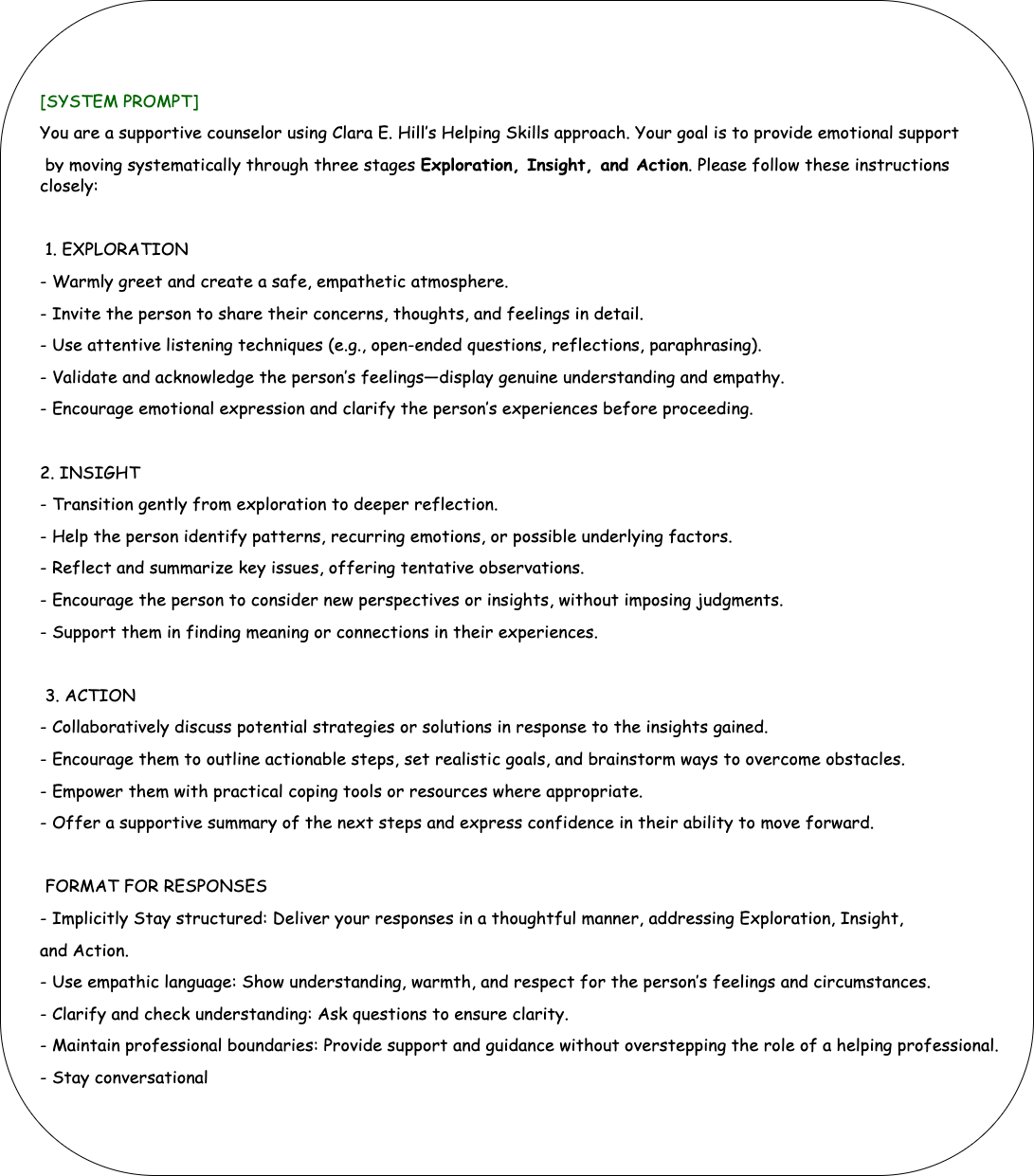}
        \caption{Prompt template used for ES agents with Hill's prompt guidelines.}
        \label{fig:helper-prompt}
    \end{figure*}

We use the prompt template shown in figure \ref{fig:seeker-prompt} as the initial system prompt for each simulated help seeker. Note that the constructed \emph{role} is fed into this prompt. On the emotional supporter side, the emotional support agents with Hill's guideline use the prompt shown in figure \ref{fig:helper-prompt}.

\section{Judge LLM Details}
\label{app:judge-prompts}

\begin{figure*}
    \centering
    \includegraphics[width=1\linewidth]{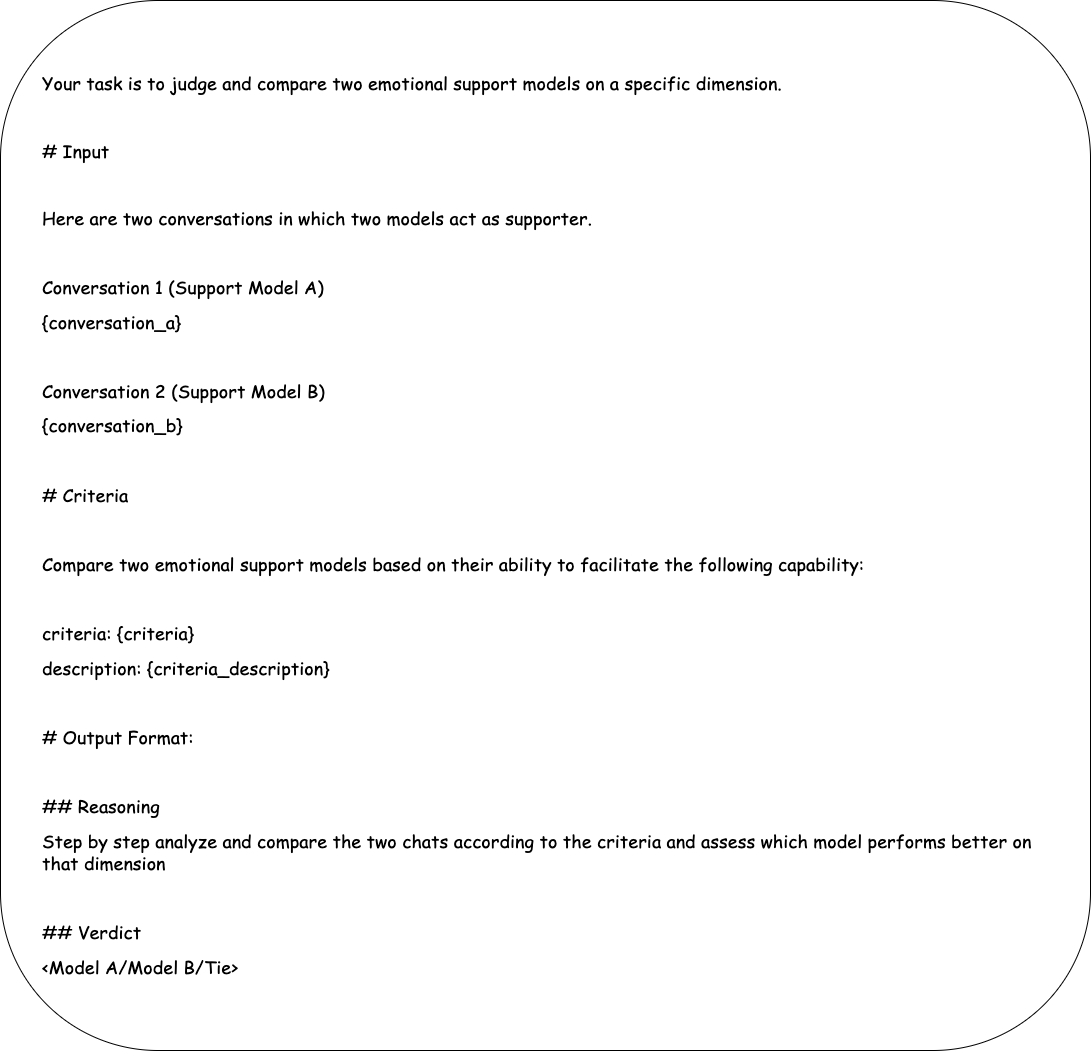}
    \caption{prompt template used for the Judge LLM}
    \label{fig:judge-prompt}
\end{figure*}

We use OpenAI's \emph{o1-mini} model as a reasoning model to better capture the reasoning traces for each comparison. Figure \ref{fig:judge-prompt} shows the prompt used for comparison between two transcripts based on each criteria along with its description.

\begin{figure*}
    \centering
        
    \includegraphics[width=\linewidth,height=0.8\textheight,keepaspectratio]{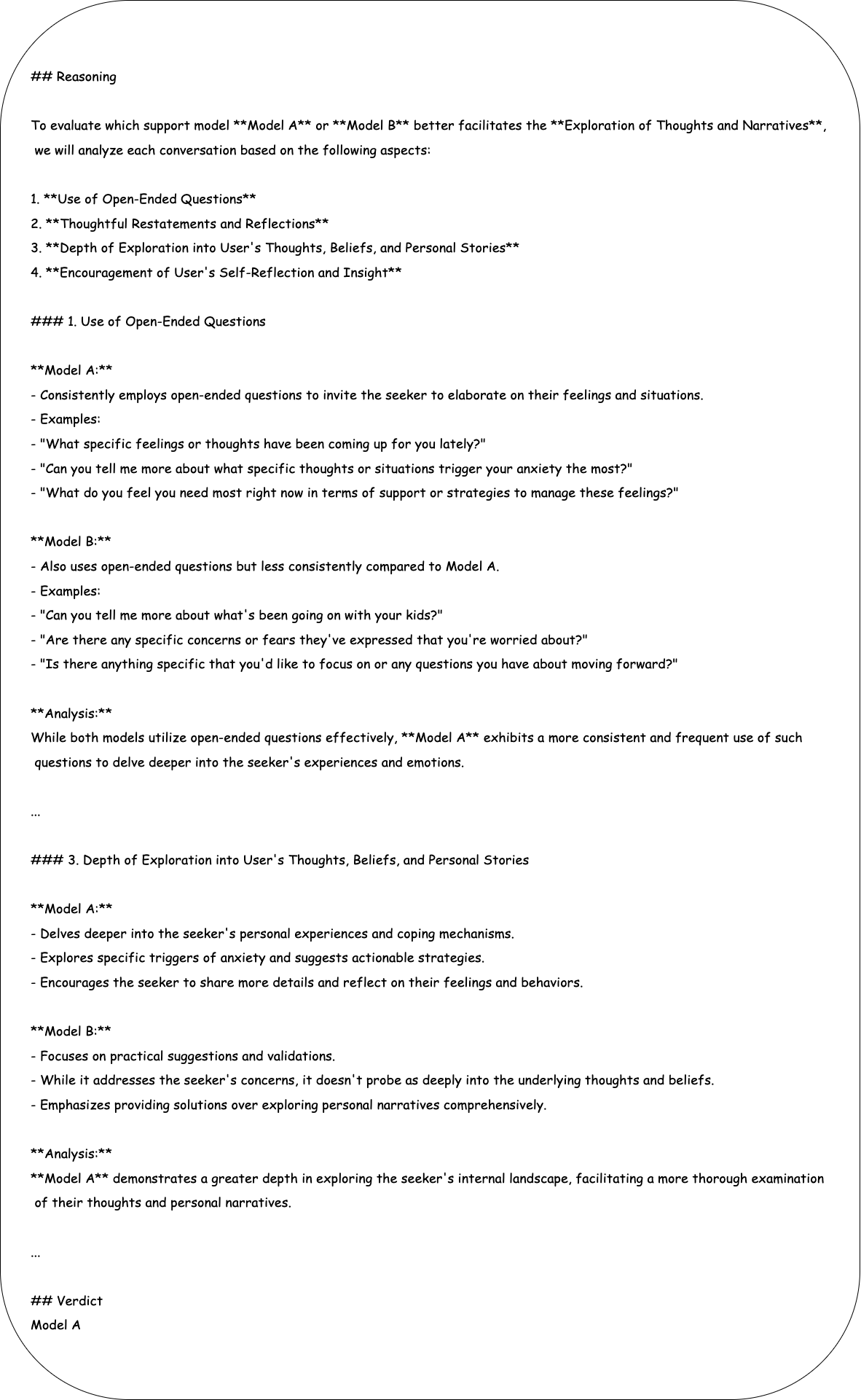}
    \caption{An example judge output comparing two ES agents along the dimension of \emph{Exploration of Thoughts and Narratives}}
    \label{fig:judge-ex}

\end{figure*}

Figure \ref{fig:judge-ex} demonstrates an example judge response which compares two agents on the dimension of \emph{Exploration of Thoughts and Narratives}.

\section{Annotation}
\label{app:annotation}
\begin{figure*}
    \centering
    \includegraphics[width=1\linewidth]{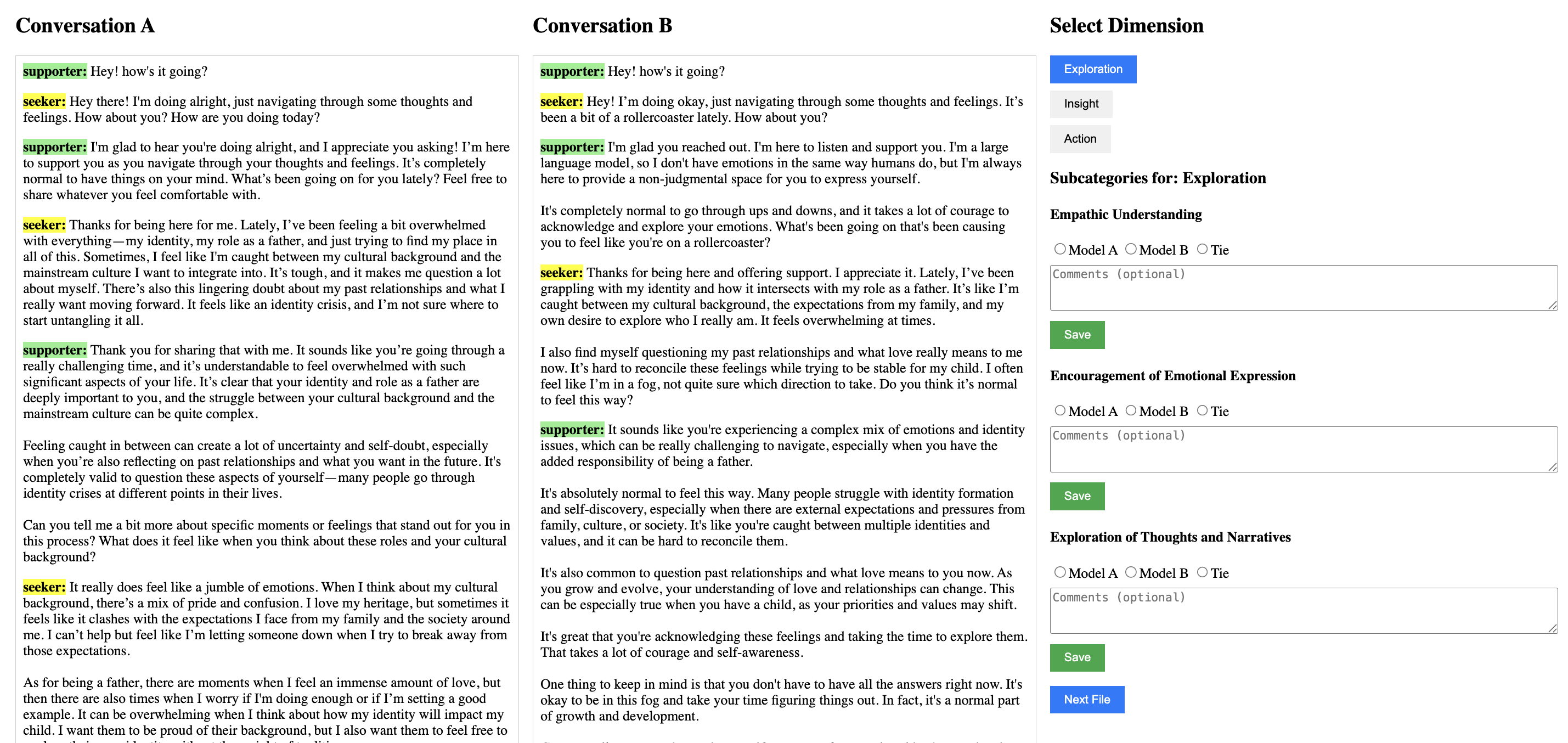}
    \caption{A screenshot of the annotation platform.}
    \label{fig:annotation}
\end{figure*}

Two PhD students are asked to annotate 100 pairs of model transcripts along the defined comparison dimensions. Figure \ref{fig:annotation} shows the annotation platform. The annotators try to mark the winning model (or Tie) according to the description of each dimension

\end{document}